\documentclass[runningheads]{llncs}

\usepackage{accv}

\usepackage{accvabbrv}

\usepackage{graphicx}
\usepackage{booktabs}
\usepackage{CJKutf8}

\usepackage[accsupp]{axessibility}  

\usepackage[hidelinks]{hyperref}

\usepackage{orcidlink}
\usepackage{multirow}

\begin{document}

\title{From Global Alignment to Local Grounding: Zero-Shot Chinese Character Recognition with Radical Verification}

\titlerunning{Zero-Shot Chinese Character Recognition with Radical Verification}


\author{
Yu-Heng Shih     \inst{1} \and
Bing-Chen Wu     \inst{1} \and
Tsz-To Wong      \inst{1} \and
Ting-En Yen      \inst{1} \and
Hong-Han Shuai   \inst{1}\orcidlink{0000-0003-2216-077X} \and
Bin-Hua Hsieh    \inst{2}\orcidlink{0009-0009-0635-0697} \and
Chien-An Chen    \inst{2} \and
Yi-Ren Yeh       \inst{3}\orcidlink{0000-0003-4264-523X} \and
Ching-Chun Huang \inst{1}\orcidlink{0000-0002-4382-5083}
}
\authorrunning{Y.-H. Shih et al.}

\institute{National Yang Ming Chiao Tung University, Taiwan\\
\email{\{ra890927.cs12,evan20010126.cs12,tsztowong.cs13,\\
tny0302.en12,hhshuai,chingchun\}@nycu.edu.tw}
\and
E.SUN Financial Holding Co., Ltd., Taiwan\\
\email{\{jeff-23314,lukechen-15953\}@esunbank.com.tw}
\and
National Kaohsiung Normal University, Taiwan\\
\email{yryeh@nknu.edu.tw}}

\maketitle

\begin{abstract}
Zero-shot Chinese character recognition (ZS-CCR) aims to recognize characters
whose categories are never observed during training, and typically relies on the
compositional structure shared between seen and unseen characters. Recent
CLIP-style methods represent this structure with the Ideographic Description
Sequence (IDS) and align it with glyph images in a shared embedding space.
However, they rely on a single global image--IDS similarity that discards the
spatial layout of radicals and, being learned only implicitly from seen classes,
generalizes poorly to unseen ones; moreover, global matching often retrieves the
correct character within the top candidates yet fails to rank it first when
characters differ only in subtle local radicals. To address these issues, we
propose a global-to-local two-stage framework. In the first stage, STG-CLIP
augments the IDS with explicit tree-position and radical-level geometric priors,
yielding a spatial-aware prototype that provides a consistent spatial
description across seen and unseen categories for high-recall global retrieval.
In the second stage, the Radical Verification Module (RVM) uses the radical instances of each retrieved candidate as queries to verify whether the corresponding radicals can be matched to spatially compatible regions in the input glyph. A margin-based gating rule activates the RVM only when the leading global candidates receive similar similarity scores. Experiments on the ICDAR2013 benchmark demonstrate that our method achieves state-of-the-art performance under the character-level zero-shot setting, obtaining 83.06\% top-1 accuracy with 2,755 seen classes. Ablation studies further show that the explicit geometric priors and radical-level verification provide complementary improvements.

\keywords{Zero-Shot Learning \and Chinese Character Recognition \and Radical Verification}
\end{abstract}

\section{Introduction}
\label{sec:intro}

Unlike alphabetic scripts, Chinese contains an enormous number of character categories, and rare characters absent from the training data are routinely encountered in everyday handwriting, historical documents, and personal or place names. Since collecting handwritten samples for every category is practically impossible, zero-shot Chinese character recognition (ZS-CCR), which recognizes characters without any handwritten training samples, provides a practical direction for handling rare and unseen characters.

In ZS-CCR, the vocabulary is split into disjoint \emph{seen} classes, which have handwritten training samples, and \emph{unseen} classes, which have none. Every class is described by its Ideographic Description Sequence (IDS), which decomposes a character into radical and structural-operator tokens; e.g., an character \begin{CJK}{UTF8}{gbsn}好\end{CJK} decomposes into \begin{CJK}{UTF8}{gbsn}女\end{CJK} and \begin{CJK}{UTF8}{gbsn}子\end{CJK} in a left--right structure, both of which also appear in seen characters. During training, the model learns to match handwritten images of seen classes to the IDS of their classes. At inference, it takes a handwritten image of an unseen class and outputs the class whose IDS best matches it. No handwritten image, glyph, radical mask, or class description of an unseen class enters optimization.

Recognition is possible because Chinese characters are strongly compositional: radicals and their spatial arrangements are shared across categories, so knowledge transfers from seen to unseen classes. Recent ZS-CCR methods adopt CLIP-style frameworks~\cite{ccr-clip, ccr-formation-tree, cross-modal-alignment} that learn a shared embedding space between glyph images and IDS sequences. Relying on a single global image--IDS similarity, however, has two limitations for fine-grained recognition.

First, the IDS is flattened into a one-dimensional sequence before text encoding, which discards the parent--child and sibling relationships of its tree topology. A prior work, FT-CLIP~\cite{ccr-formation-tree}, encodes only the parent relation of each node. More importantly, existing methods lack explicit geometric descriptors of where each radical resides and how much space it occupies, so the model must implicitly infer these spatial distributions from seen characters. When transferred to unseen characters with novel radical combinations and layouts, the implicitly learned positional prior extrapolates poorly, causing a seen-to-unseen spatial generalization gap.

Second, a single global similarity score is frequently insufficient for fine-grained discrimination. Characters with similar overall layouts often differ by only a single local radical. Global alignment effectively retrieves the correct category within the top-$k$ candidates but struggles to rank it first. Global matching thus suits candidate retrieval but not the final decision.

To address these limitations, we propose a coarse-to-fine two-stage recognition framework grounded in a dual-role interpretation of the IDS: the IDS is encoded into a spatial-aware class prototype for global retrieval, while the same decomposition serves as radical-level hypotheses to be verified locally. Rather than leaving radical positions to be inferred from data, the first stage, Spatial Tree-Geometric CLIP (STG-CLIP), injects tree-position embeddings (depth and sibling indices) and canonical radical geometric priors (normalized centroid, bounding box, and area ratio) into the IDS representation as explicit signals. Because these priors are derived from canonical glyphs and remain independent of category visibility, STG-CLIP establishes a consistent spatial description across seen and unseen categories; the resulting prototypes are used for global image--prototype alignment to retrieve top-$k$ candidates at high recall. The second stage, the Radical Verification Module (RVM), treats the IDS decomposition of each retrieved candidate as a set of radical hypotheses. Through cross-attention between radical queries and localized image features, the RVM extracts radical-level visual evidence and evaluates each candidate by radical presence and glyph coverage. A margin-based gating rule invokes the RVM only when the leading global candidates are indistinguishable.

Our main contributions are summarized as follows:
\begin{itemize}
\item{\textbf{Spatial-Aware IDS Representation}}
We introduce a tree- and geometry-aware IDS representation that injects depth and sibling embeddings alongside canonical geometry (centroid, bounding box, and area ratio) into IDS tokens. This explicit class-side information mitigates the seen-to-unseen spatial gap.

\item{\textbf{Coarse-to-Fine Framework with Margin Gating}}
We develop a margin-gated coarse-to-fine inference framework that separates global candidate retrieval from local verification, triggering verification only for ambiguous decisions to balance efficiency and accuracy.

\item{\textbf{Radical Verification Module (RVM)}}
The RVM advances ZS-CCR from global similarity matching to candidate-conditioned local verification. By evaluating radical presence and glyph coverage, the RVM leverages localized visual evidence to resolve subtle ambiguities among visually similar characters.
\end{itemize}

\section{Related Work}
\label{sec:related}

\subsection{Contrastive Learning-based Methods for ZS-CCR}
\label{subsec:clip}

Contrastive learning frameworks, particularly CLIP-inspired methods \cite{clip2021}, show strong potential in ZS-CCR by aligning visual features with class-level semantic text descriptions. CCR-CLIP \cite{ccr-clip} pioneers this paradigm by matching printed images with their 1D Ideographic Description Sequences (IDS) in a shared space. However, flattening the 2D spatial layout into a 1D sequence discards the hierarchical tree topology. To capture this, FT-CLIP \cite{ccr-formation-tree} introduces a tree encoder operating on a novel Formation Tree, while Cai and Zhu \cite{cross-modal-alignment} align both local components and global features to enrich cross-modal representations.

Although these methods improve structural modeling via symbolic or relative component relationships, they lack absolute geometric awareness. They fail to associate radical instances with continuous, character-dependent geometric attributes (e.g., centroid, bounding-box size, and area ratio). Conversely, our STG-CLIP augments tokens with explicit tree-position information and canonical geometry, while separating global retrieval from candidate-specific verification.

\subsection{Generative Sequence Recognition Methods}
\label{subsec:generative}

Another paradigm treats ZS-CCR as an image-to-sequence translation task. Methods like DenseRAN \cite{denseRAN} and SD/SLD \cite{sd-sld} employ visual encoders with auto-regressive (AR) decoders to sequentially generate radical or stroke sequences, enabling the recognition of unseen classes via reusable components. However, AR decoding is prone to error accumulation, where early errors propagate through the sequence, and requires complex post-processing to parse valid labels. In contrast, our method bypasses sequence generation entirely, treating the IDS as established class-side information to directly retrieve character classes through image--prototype alignment.

\subsection{Radical Detection and Localization Methods}
\label{subsec:localization}

To enhance spatial grounding, several studies formulate radical recognition as a detection or localization problem. JRED \cite{JRED} employs learnable radical embeddings as sliding detectors over multi-scale feature maps to generate explicit response maps. Similarly, LERRNet \cite{lerrnet} introduces attribute hint vectors initialized via Word2Vec to guide the decoder toward specific radical regions, proving that local evidence effectively complements holistic representations. Nevertheless, these end-to-end models predict component sequences directly rather than verifying candidate hypotheses. Our Radical Verification Module (RVM) instead acts as a candidate-conditioned verifier, extracting localized visual evidence to resolve global prediction ambiguities.

\subsection{Radical Representation Learning}
\label{subsec:radical-representation}

Representing individual radical identities inherently before structural aggregation is also critical. Early methods represent radicals using discrete multi-hot or random encodings, ignoring visual properties and distorting the encoding space topology. To alleviate this, FaRE \cite{fareZhan2024} introduces a feature-aware encoding strategy that extracts implicit visual features from rendered font images. While FaRE improves radical category embeddings, existing IDS-based methods still treat radicals as static semantic tokens, ignoring their character-dependent location, scale, or aspect ratio. Since the same radical exhibits distinct physical variations across different characters, our method models the role of a radical *instance* by attaching canonical geometric attributes to the corresponding IDS tokens, resolving their spatial blindness.

\section{Proposed Method}
\label{sec:methods}

\begin{figure}
    \centering
    \includegraphics[width=1.0\linewidth]{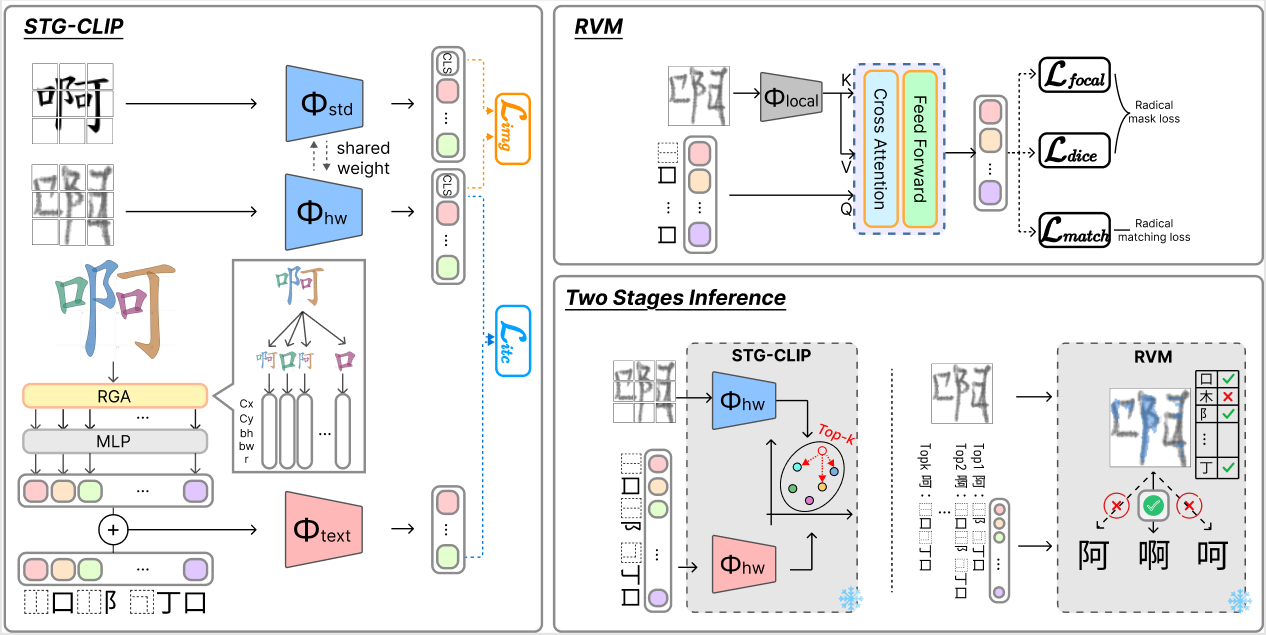}
    \caption{Overview of our framework. \textbf{STG-CLIP}~(Left): the first stage learns a shared image--text space by constructing tree- and geometry-aware IDS prototypes. \textbf{RVM}~(Top-right): the RVM uses the radical occurrences of each candidate as queries to verify whether they are grounded in the input image. \textbf{Two-stage Inference}~(Bottom-right): STG-CLIP first retrieves the Top-$K$ candidates by global similarity; the RVM then verifies the radicals of each candidate against the input glyph and selects the final prediction when the leading candidates are globally ambiguous.}
    \label{fig:pipeline}
\end{figure}

\subsection{Problem Formulation and Framework Overview}
\label{sec:problem-formulation}

In zero-shot Chinese character recognition (ZS-CCR), the objective is to recognize handwritten character instances from categories that are not observed during training. Let $\mathcal{Y}_{\text{seen}}$ and $\mathcal{Y}_{\text{unseen}}$ denote the sets of seen and unseen character categories, respectively, where $\mathcal{Y}_{\text{seen}} \cap \mathcal{Y}_{\text{unseen}} = \emptyset$. During training, handwritten character images are available only for categories in $\mathcal{Y}_{\text{seen}}$. During inference, given a handwritten query image $I^*$ from an unseen category, the model predicts its label $\hat{y} \in \mathcal{Y}_{\text{unseen}}$. 

In our setting, IDS descriptions and canonical geometric descriptors are treated as class-side side information available for all categories, whereas printed glyph images and radical masks belonging to unseen classes are strictly excluded from model optimization.

Following the dual-role interpretation of IDS, the proposed framework separates global candidate retrieval from local radical verification, as illustrated in Figure~\ref{fig:pipeline}. The first stage, Spatial Tree-Geometric CLIP (\textbf{STG-CLIP}), augments IDS tokens with explicit tree-position and radical-geometry information to learn a shared image--prototype embedding space. Given a handwritten query, STG-CLIP performs global alignment to retrieve a compact top-$K$ candidate shortlist $\mathcal{C}_K \subset \mathcal{Y}_{\text{unseen}}$ at high recall. 

The second stage, the Radical Verification Module (\textbf{RVM}), treats the IDS decomposition of each retrieved candidate as a specific set of radical hypotheses. Through cross-attention between radical queries and localized visual features, the RVM evaluates whether the queried radicals are visually present and whether their predicted regions spatially support the observed glyph foreground. Finally, a margin-based gating rule decides whether to invoke this local verification to re-rank the candidates or safely trust the global top-1 prediction.

\subsection{Tree- and Geometry-Aware IDS Representation}
\label{sec:ids-representation}

For each character category $y$ (either seen or unseen), we construct a geometrically augmented IDS representation $S_{\text{geo}}(y)$ and encode it into a class prototype $\mathbf{t}_y$. This representation enriches the original IDS token identities by injecting lightweight tree-position information and canonical radical-level geometry.

\subsubsection{IDS Tree-Position Encoding}

The text encoder in CLIP-style frameworks is inherently designed for sequential inputs, meaning its self-attention mechanism does not presuppose any hierarchical tree structures. When the IDS tree is flattened into a one-dimensional sequence, standard sequential encoders struggle to distinguish sibling components at the same hierarchical level from nested components across different levels. 

To preserve the structural relations implicit in the tree topology, we parse the IDS into a tree structure and assign each token a depth index $d_i$ and a sibling index $s_i$. The depth index denotes the token's distance from the root node, while the sibling index captures the left-to-right order among nodes sharing the same parent. These attributes provide lightweight hierarchical cues to guide the sequence encoding.

\subsubsection{Radical Geometry Augmentation}

While the tree-position encoding captures the symbolic organization of the IDS tree, it lacks absolute physical awareness regarding where each component resides and how much space it occupies. To address this spatial blindness, we augment the tokens with continuous geometric descriptors computed from canonical printed glyph masks. 

For a radical token, its corresponding stroke paths are grouped and rendered onto a binary canvas to obtain its component mask. For a structural operator token, its geometric attribute is derived from the spatial aggregation (the enclosing bounding box and union of masks) of its corresponding subtree. From the resulting mask of each token $i$, we extract a five-dimensional geometric prior vector:
\begin{equation}
\mathbf{g}_i = [c_x, c_y, b_w, b_h, r_{\text{area}}] \in \mathbb{R}^{5},
\label{eq:geometry-feature}
\end{equation}
where $(c_x,c_y)$ is the normalized centroid within the character canvas, $(b_w,b_h)$ represents the normalized bounding-box width and height, and $r_{\text{area}}$ denotes the foreground area ratio computed specifically within its own bounding box rather than the whole glyph plane. These descriptors serve as stable class-level spatial priors independent of handwriting styles.

Given the original token embedding $\mathbf{x}_i$, we form a fused token representation by adding the tree-position and projected geometric embeddings:
\begin{equation}
\mathbf{x}_i^{\text{geo}} = \mathbf{x}_i + E_{\text{depth}}(d_i) + E_{\text{sibling}}(s_i) + \text{MLP}(\mathbf{g}_i),
\label{eq:geometry-token}
\end{equation}
where $E_{\text{depth}}(\cdot)$ and $E_{\text{sibling}}(\cdot)$ are learnable tree-position embeddings, and $\text{MLP}(\cdot)$ projects the geometric descriptor into the token embedding dimension. The final augmented sequence for category $y$ is denoted as $S_{\text{geo}}(y) = \{\mathbf{x}_i^{\text{geo}}\}_{i=1}^{L_y}$, where $L_y$ is the sequence length.

\subsubsection{IDS Prototype Encoding}

The spatial-aware token sequence is fed into a text encoder $\Psi(\cdot)$ to generate the final class prototype vector:
\begin{equation}
\mathbf{t}_y = \Psi(S_{\text{geo}}(y)).
\label{eq:text-prototype}
\end{equation}
When instantiated with a CLIP-style Transformer, the hidden state at the end-of-sequence (EOS) token is projected into the shared embedding space as $\mathbf{t}_y$. Notably, this representation serves a dual purpose: $\mathbf{t}_y$ acts as a global retrieval target in STG-CLIP, while the underlying symbolic radical decomposition is passed to the RVM as structural hypotheses for local verification.

\subsection{Global Image--Prototype Alignment (STG-CLIP)}
\label{sec:stg-clip}

The first stage, \textbf{STG-CLIP}, learns a shared embedding space between handwritten character images and the spatially enriched IDS prototypes to perform coarse retrieval.

\subsubsection{Image and Prototype Representation}

Given a handwritten image $I_{\text{hw}}$ and its corresponding canonical printed glyph $I_{\text{std}}$ from a seen category, a shared image encoder $\Phi_{\text{img}}(\cdot)$ extracts their visual embeddings:
\begin{equation}
\mathbf{v}_{\text{hw}} = \Phi_{\text{img}}(I_{\text{hw}}), \qquad \mathbf{v}_{\text{std}} = \Phi_{\text{img}}(I_{\text{std}}).
\label{eq:image-embedding}
\end{equation}
The printed-glyph branch acts purely as an auxiliary regularizer during training and is completely omitted at inference. All visual and text embeddings are $L_2$-normalized before calculating similarity scores.

\subsubsection{Contrastive Alignment with Style Regularization}

A symmetric InfoNCE objective is then applied to the distinct image--prototype pairs $\{(\bar{\mathbf{v}}_c,\mathbf{t}_c)\}_{c=1}^{C}$ within the mini-batch:
\begin{equation}
\mathcal{L}_{\text{itc}} = -\frac{1}{2C} \sum_{c=1}^{C} \left[ \log \frac{ \exp(\text{sim}(\bar{\mathbf{v}}_c,\mathbf{t}_c)/\tau) }{ \sum_{j=1}^{C} \exp(\text{sim}(\bar{\mathbf{v}}_c,\mathbf{t}_j)/\tau) } + \log \frac{ \exp(\text{sim}(\mathbf{t}_c,\bar{\mathbf{v}}_c)/\tau) }{ \sum_{j=1}^{C} \exp(\text{sim}(\mathbf{t}_c,\bar{\mathbf{v}}_j)/\tau) } \right],
\label{eq:contrastive-loss}
\end{equation}
where $\text{sim}(\cdot,\cdot)$ denotes cosine similarity and $\tau$ is a temperature parameter.

To mitigate the visual domain gap caused by diverse handwriting styles, we introduce an auxiliary style regularization term that encourages the handwritten embedding to align closely with the corresponding canonical printed glyph:
\begin{equation}
\mathcal{L}_{\text{style}} = \frac{1}{B} \sum_{i=1}^{B} \left[ 1 - \text{sim}(\mathbf{v}_{\text{hw},i}, \mathbf{v}_{\text{std},i}) \right].
\label{eq:style-loss}
\end{equation}
The overall objective for global alignment is formulated as:
\begin{equation}
\mathcal{L}_{\text{global}} = \mathcal{L}_{\text{itc}} + \lambda_{\text{style}} \mathcal{L}_{\text{style}},
\label{eq:global-loss}
\end{equation}
where $\lambda_{\text{style}}$ controls the weight of the style regularization term.

\subsubsection{Top-$K$ Candidate Retrieval}

During inference, given a query handwritten image $I^*$, its visual feature vector $\mathbf{v}^* = \Phi_{\text{img}}(I^*)$ is matched against all possible character categories in the entire recognition vocabulary to compute global similarity scores:
\begin{equation}
s_{\text{global}}(y) = \text{sim}(\mathbf{v}^*, \mathbf{t}_y), \qquad y \in \mathcal{Y}_{\text{vocabulary}}.
\label{eq:global-score}
\end{equation}
The coarse shortlist of candidates is then retrieved by selecting the top-$K$ scoring classes:
\begin{equation}
\mathcal{C}_K = \text{TopK}_{y \in \mathcal{Y}_{\text{vocabulary}}}\; s_{\text{global}}(y).
\label{eq:topk}
\end{equation}

\subsection{Radical Verification Module}
\label{sec:rvm}

To resolve fine-grained ambiguities among visually similar characters that share a nearly identical global layout but differ by a single radical, we introduce the \textbf{Radical Verification Module (RVM)}. The RVM shifts the focus from holistic semantic matching to localized structural alignment, explicitly verifying whether a candidate's radicals are visually grounded in the input image.

\subsubsection{Localized Visual Features}

To extract rich local details, a separate CNN backbone $\Phi_{\text{local}}(\cdot)$—specifically ResNet-50~\cite{resnet}—is dedicated to the RVM and trained independently of STG-CLIP. The generated visual feature maps are flattened and projected into a sequence of patch tokens $F = \{\mathbf{f}_1, \dots, \mathbf{f}_N\} \in \mathbb{R}^{N \times d}$. Standard 2D sinusoidal position embeddings are added to $F$ to preserve the spatial layout of the character.

\subsubsection{Decoder-based Structural Alignment}

For an evaluated candidate $y$, its IDS decomposition yields an ordered sequence of radical occurrences $\mathcal{R}(y) = (r_1, \dots, r_{Q_y})$. Each radical occurrence is initialized as a learnable embedding to serve as a decoder query:
\begin{equation}
R^{(0)} = \left( E_{\text{rad}}(r_1), \dots, E_{\text{rad}}(r_{Q_y}) \right) \in \mathbb{R}^{Q_y \times d}.
\label{eq:radical-query}
\end{equation}
Unlike STG-CLIP text representations, these queries do not encode canonical geometric priors, as handwritten and printed radicals are not pixel-aligned.

These queries are fed into a cascade of $L$ transformer decoder layers to attend to the localized visual features $F$. At the $l$-th layer, the representations are refined via cross-attention:
\begin{equation}
\widetilde{R}^{(l)} = \text{LayerNorm}\left( R^{(l-1)} + \text{CrossAttn}\left(Q=R^{(l-1)}, K=F, V=F\right) \right),
\label{eq:rvm-cross-attn}
\end{equation}
\begin{equation}
R^{(l)} = \text{LayerNorm}\left( \widetilde{R}^{(l)} + \text{FFN}\left(\widetilde{R}^{(l)}\right) \right).
\label{eq:rvm-ffn}
\end{equation}

The terminal contextualized radical representations $R^{(L)} \in \mathbb{R}^{Q_y \times d}$ are subsequently forwarded to two parallel prediction heads. The match head applies a linear projection followed by a sigmoid function to output an existence probability $\hat{s}_q \in [0,1]$ for each radical query, quantifying whether the component is visually supported:
\begin{equation}
\hat{s}_q = \sigma\left( h_{\text{match}}\left( R_q^{(L)} \right) \right).
\label{eq:match-score}
\end{equation}
Simultaneously, the mask head projects each token into a spatial response map, which is reshaped and bilinearly upsampled to the input resolution, yielding a soft attention mask $\hat{M}_q \in [0,1]^{H \times W}$ that dynamically delineates the predicted location of the $q$-th radical:
\begin{equation}
\hat{M}_q = h_{\text{mask}}\left( R_q^{(L)} \right).
\label{eq:mask-output}
\end{equation}

\subsubsection{Multi-Task Optimization}

During training, positive queries are formed by the radicals inherently present in the character, while negative queries are randomly sampled from absent categories. 

Radical existence is supervised via a binary cross-entropy loss:
\begin{equation}
\mathcal{L}_{\text{match}} = -\frac{1}{Q}\sum_{q=1}^{Q} \left[ s_q \log(\hat{s}_q) + (1-s_q)\log(1-\hat{s}_q) \right],
\label{eq:match-loss}
\end{equation}
where $s_q \in \{0,1\}$ represents the ground-truth presence.

For spatial localization, since radical strokes are extremely sparse on the image lattice, we optimize positive queries using a combination of Focal loss (to suppress easy background pixels) and Dice loss (to optimize region overlap):
\begin{equation}
\mathcal{L}_{\text{focal}}^{(q)} = -\frac{1}{HW}\sum_{i=1}^{HW} \alpha_t (1 - p_{q,i})^{\gamma} \log(p_{q,i}),
\label{eq:focal-loss}
\end{equation}
where $p_{q,i} = \hat{m}_{q,i}$ if the ground-truth mask pixel $m_{q,i} = 1$, and $p_{q,i} = 1 - \hat{m}_{q,i}$ otherwise; $\alpha_t$ balances foreground/background pixel distributions. The Dice loss is defined as:
\begin{equation}
\mathcal{L}_{\text{dice}}^{(q)} = 1 - \frac{2 \langle \hat{M}_q, M_q \rangle + \epsilon}{\|\hat{M}_q\|_1 + \|M_q\|_1 + \epsilon},
\label{eq:dice-loss}
\end{equation}
where $\langle \cdot, \cdot \rangle$ denotes the inner product and $\epsilon$ is a small constant. Negative queries bypass the spatial losses. The total multi-task objective is:
\begin{equation}
\mathcal{L}_{\text{RVM}} = \mathcal{L}_{\text{match}} + \lambda_1 \frac{1}{|\mathcal{Q}^+|} \sum_{q \in \mathcal{Q}^+} \mathcal{L}_{\text{focal}}^{(q)} + \lambda_2 \frac{1}{|\mathcal{Q}^+|} \sum_{q \in \mathcal{Q}^+} \mathcal{L}_{\text{dice}}^{(q)},
\label{eq:rvm-loss}
\end{equation}
where $\mathcal{Q}^+$ is the set of positive queries, and $\lambda_1, \lambda_2$ are balancing weights.

\subsection{Training and Inference Pipeline}
\label{sec:pipeline-paradigm}

To ensure robust radical grounding and efficient deployment, we design a two-stage training strategy followed by a gated inference mechanism.

\paragraph{Two-Stage Training Strategy} 
The training procedure consists of a printed pre-training phase followed by a handwritten fine-tuning phase. In the printed pre-training stage (strong supervision), the RVM is optimized on standard, clean printed glyphs from the seen character classes. Because printed glyphs offer precise, pixel-level radical mask annotations, all three loss functions ($\mathcal{L}_{\text{match}}$, $\mathcal{L}_{\text{focal}}$, and $\mathcal{L}_{\text{dice}}$) are fully activated. This stage forces the RVM queries to firmly anchor onto their corresponding visual regions and layouts. In the subsequent handwritten fine-tuning stage (weak supervision), the model transitions to real-world handwritten images from the seen classes. Since real handwritten data lacks radical-level pixel annotations, the mask head is frozen to preserve the anchoring capabilities learned from printed glyphs. The remaining parts of the module are optimized using exclusively the character-level occurrence loss $\mathcal{L}_{\text{match}}$, allowing the local backbone and cross-attention blocks to adapt to handwritten deformations without requiring dense spatial labels.

\paragraph{Gated Coarse-to-Fine Inference} 
At inference, the RVM serves as a plug-and-play verifier for the retrieved candidate shortlist $\mathcal{C}_K$. For each candidate $y_k \in \mathcal{C}_K$, its constituent radicals are queried to output existence scores $\hat{s}_j$ and soft masks $\hat{M}_j$. We evaluate each candidate through a two-fold local verification score. The first component is radical presence ($s_{\text{pres}}$), which computes the average existence probability of all queried radicals to penalize extraneous components:
\begin{equation}
s_{\text{pres}}(y_k) = \frac{1}{|\mathcal{R}(y_k)|} \sum_{j=1}^{|\mathcal{R}(y_k)|} \hat{s}_j.
\label{eq:presence-score}
\end{equation}
The second component is glyph coverage ($s_{\text{cov}}$), which measures whether the union of the predicted radical masks covers the actual observed strokes in the handwritten foreground mask $M_{\text{fg}}$ obtained by simple image binarization:
\begin{equation}
M_{\text{union}}(y_k) = \bigcup_{j=1}^{|\mathcal{R}(y_k)|} \mathbb{I} \left( \hat{M}_j > \theta \right),
\label{eq:mask-union}
\end{equation}
\begin{equation}
s_{\text{cov}}(y_k) = \frac{\left\| M_{\text{union}}(y_k) \cap M_{\text{fg}} \right\|_1}{\left\| M_{\text{fg}} \right\|_1 + \epsilon},
\label{eq:coverage-score}
\end{equation}
where $\mathbb{I}(\cdot)$ binarizes the soft masks at threshold $\theta$. The joint local verification score is formulated as:
\begin{equation}
s_{\text{local}}(y_k) = s_{\text{pres}}(y_k) \cdot s_{\text{cov}}(y_k).
\label{eq:local-score}
\end{equation}
To maintain retrieval efficiency and avoid unnecessary computations, a margin-based gating rule dictates the final prediction. Let $y_{(1)}$ and $y_{(2)}$ be the top-1 and top-2 candidates ranked by the first-stage global score $s_{\text{global}}$. We calculate the confidence margin:
\begin{equation}
\Delta = s_{\text{global}}(y_{(1)}) - s_{\text{global}}(y_{(2)}).
\label{eq:global-margin}
\end{equation}
The final predicted label $\hat{y}$ is determined by:
\begin{equation}
\hat{y} =
\begin{cases}
\displaystyle \arg\max_{y_k \in \mathcal{C}_K} s_{\text{local}}(y_k), & \text{if } \Delta \le \delta, \\[8pt]
y_{(1)}, & \text{otherwise},
\end{cases}
\label{eq:final-prediction}
\end{equation}
where $\delta$ is a predefined threshold tuned on the validation set. When the global confidence margin $\Delta$ is large ($\Delta > \delta$), the global top-1 prediction is directly trusted for maximum efficiency. Conversely, when the top two candidates are neck-and-neck ($\Delta \le \delta$), indicating a fine-grained structural ambiguity, the gate is activated, and the local verification score $s_{\text{local}}$ is leveraged to resolve the ambiguity based on explicit local evidence.

\section{Experiments}

\subsection{Dataset and Implementation Details}

\subsubsection{Dataset.}

We conduct experiments on two offline handwritten Chinese character
datasets, CASIA-HWDB1.0-1.1~\cite{CASIA-HWDB} and ICDAR2013~\cite{ICDAR2013}.
CASIA-HWDB1.0-1.1 aggregates 2,678,424 isolated character samples over
3,881 classes, among which 3,755 belong to the level-1 commonly used
character set of GB2312-80, and we use these samples for training.
ICDAR2013 is adopted as the test set; it contains 224,419 handwritten
samples collected from 60 writers that are disjoint from those in the
training set, while covering the identical 3,755 level-1 categories.

\label{sec:dataset-rvm-gt}
To supervise the RVM at the radical level, we additionally construct
ground-truth radical masks from printed glyphs, since handwritten samples
provide no pixel-level radical annotations. Specifically, we render characters
from makemeahanzi and, following the IDS decomposition, group the strokes
belonging to each radical and rasterize them onto the glyph canvas to obtain a
binary mask for that radical. These printed-glyph masks are used only in the
RVM pre-training stage; during handwritten fine-tuning, the mask head is frozen
and no mask annotations are required. Illustrative examples are provided in the
supplementary material.

\subsubsection{Implementation Details.}

Our framework is implemented in PyTorch and evaluated across two processing granularities. For the global retrieval stage, the STG-CLIP image encoder operates on $32\times32$ inputs, and the style-regularization weight is set to $\lambda_{\text{style}}{=}1$. For the local verification stage, the Radical Verification Module (RVM) adopts a dedicated backbone operating on $64\times64$ inputs, and the balancing weights for multi-task optimization are set to $\lambda_1{=}5$ and $\lambda_2{=}1$. At inference time, the gating threshold $\delta$ is adaptively selected per setting on the validation set. All remaining hardware configurations, training objectives, and optimization hyperparameters are deferred to the Supplementary Material.

\begin{table*}[t]
\centering
\caption{Character-level and radical-level zero-shot recognition accuracy on handwritten datasets under different numbers of seen training classes.}
\label{tab:zeroshot}
\resizebox{\textwidth}{!}{%
\begin{tabular}{l ccccc ccccc}
\toprule
& \multicolumn{5}{c}{Character Zero-shot} & \multicolumn{5}{c}{Radical Zero-shot} \\
\cmidrule(lr){2-6} \cmidrule(lr){7-11}
Method & 500 & 1000 & 1500 & 2000 & 2755 & 50 & 40 & 30 & 20 & 10 \\
\midrule
DenseRAN~\cite{denseRAN}       & 1.70\%  & 8.44\%  & 14.71\% & 19.51\% & 30.68\% & 0.21\%  & 0.29\%  & 0.25\%  & 0.42\%  & 0.69\%  \\
HDE~\cite{hde}                 & 4.90\%  & 12.77\% & 19.25\% & 25.13\% & 33.49\% & 3.26\%  & 4.29\%  & 6.33\%  & 7.64\%  & 9.33\%  \\
SD~\cite{sd-sld}               & 5.60\%  & 13.85\% & 22.88\% & 25.73\% & 37.91\% & 5.28\%  & 6.87\%  & 9.02\%  & 14.67\% & 15.83\% \\
STAR~\cite{star}               & 7.54\%  & 19.47\% & 27.79\% & 35.53\% & 43.86\% & 6.95\%  & 12.28\% & 14.74\% & 18.37\% & 23.23\% \\
FaRE~\cite{fareZhan2024}       & 7.21\%  & 21.78\% & 36.58\% & 47.33\% & 57.17\% &  -       & -       & -       & -       & -       \\
JRED~\cite{JRED}               & 25.19\%  & 36.21\% & 41.93\% & 42.05\% & 56.26\% & 10.21\% &  8.30\% & 12.67\% & 19.13\% & 27.08\% \\
CCR-CLIP~\cite{ccr-clip}        & 21.79\% & 42.99\% & 55.86\% & 62.99\% & 72.98\% & 11.15\% & 13.85\% & 16.01\% & 16.76\% & 15.96\% \\
LERRNet~\cite{lerrnet}        & \textbf{32.73\%} & \textbf{55.36\%} & \textbf{66.01\%} & 70.81\% & 73.59\% &  8.35\% & 11.22\% & 14.81\% & 15.98\% & 19.37\% \\
FT-CLIP~\cite{ccr-formation-tree}         & 26.17\% & 50.68\% & 62.25\% & 70.39\% & 78.96\% & \textbf{12.34\%} & \textbf{16.66\%} & 21.51\% & 24.65\% & 27.80\% \\
\cmidrule(lr){1-11}
\textbf{STG-CLIP}              & 21.25\% & 39.38\% & 59.91\% & 69.96\% & 82.26\% &  8.89\% & 14.26\% & \textbf{21.62\%} & \textbf{27.63\%} & 37.30\% \\
\textbf{STG-CLIP + RVM}        & 22.12\% & 40.52\% & 63.93\% & \textbf{71.48\%} & \textbf{83.06\%} & 8.74\% & 14.22\% & 21.44\% & 27.52\% & \textbf{37.76\%} \\
\bottomrule
\end{tabular}%
}
\end{table*}

\subsection{Experiments on Unseen Character Recognition}

Zero-shot learning refers to the ability to generalize to unseen categories. In Chinese character recognition, this setting is particularly important because many characters may be absent from the training set, while their structures can still be described by IDS. We conduct experiments under two zero-shot settings, including character zero-shot and radical zero-shot. During testing, all categories appearing in the corresponding experiment are used as candidates.

\subsubsection{Character Zero-shot Settings.}

The selection of categories in the training and testing sets has a significant impact on zero-shot Chinese character recognition. For handwritten characters, we first sort the 3755 level-1 commonly used Chinese characters according to the order of GB2312. Samples from CASIA-HWDB whose labels belong to the first $m$ characters are used as the training set, where $m \in \{500, 1000, 1500, 2000, 2755\}$. Samples from ICDAR2013 whose labels belong to the last 1000 characters are used as the test set.

\subsubsection{Radical Zero-shot Settings.}

In the radical zero-shot setting, characters are split according to the frequency of their radicals. A character is assigned to the test set if it contains at least one radical whose frequency is lower than a threshold $n$; otherwise, it is assigned to the training set. We evaluate five thresholds, where $n \in \{50, 40, 30, 20, 10\}$. This setting evaluates whether the model can generalize to characters containing rare radical compositions.

\subsection{Comparison with Existing Methods}

Table~\ref{tab:zeroshot} reports accuracy under the character and radical zero-shot settings. CLIP-based methods clearly outperform radical-based ones, and among them our method achieves the best results on most settings. On the character setting, STG-CLIP+RVM reaches 83.06\% at 2755 seen classes, surpassing the previous best FT-CLIP by 4.10 points and CCR-CLIP by 10.08. RVM consistently improves STG-CLIP across all seen-class counts, most notably by 4.02 points at 1500, confirming that local radical verification resolves the fine-grained ambiguities left by the global stage. Under the radical setting, however, RVM yields little gain: since this split partitions data by radicals, the test characters contain radicals unseen during training, which RVM cannot reliably verify.

STG-CLIP is less competitive under small training data, trailing CCR-CLIP at 500 and 1000 seen classes. We attribute this to our ViT-based image encoder, which lacks the local inductive biases of CCR-CLIP's convolutional encoder and thus needs more data to train well. The same crossover appears in both settings at a comparable data scale, indicating that ViT is more data-hungry but scales better once data is sufficient.

\subsection{Ablation Studies}

\begin{table}[t]
\centering
\caption{Cumulative ablation study of the proposed components under the character zero-shot setting with 2,755 seen classes. Each row adds one component to the previous setting.}
\label{tab:ablation}
\small
\setlength{\tabcolsep}{10pt}
\renewcommand{\arraystretch}{0.95}
\begin{tabular}{lcc}
\toprule
Cumulative setting & Top-1 Acc. & $\Delta$ \\
\midrule
Baseline & 76.27 & -- \\
+ IDS Prototype Encoding & 77.78 & +1.51 \\
+ Radical Geometry Augmentation & 82.26 & +4.48 \\
+ RVM & 83.06 & +0.80 \\
\bottomrule
\end{tabular}
\end{table}


Table~\ref{tab:ablation} evaluates the cumulative impact of our components. Compared to a plain CLIP baseline ($76.27\%$), reinjecting tree topology (depth and sibling indices) lifts accuracy to $77.78\%$ ($+1.51\%$). Further augmenting tokens with explicit canonical geometry yields the most substantial improvement to $82.26\%$ ($+4.48\%$), suggesting that explicit physical attributes mitigate the seen-to-unseen spatial gap. Lastly, incorporating the RVM provides an additional boost to $83.06\%$ ($+0.80\%$), demonstrating that localized structural verification resolves fine-grained global ambiguities. Note that per-component gains depend on the order of addition, and this ablation does not fully separate the encoding design from the added geometric information.

\subsubsection{Selection of $K$ and $\delta$}
\label{sec:ablation_selection_k_delta}
\begin{figure}[t]
\centering
\includegraphics[width=0.50\linewidth]{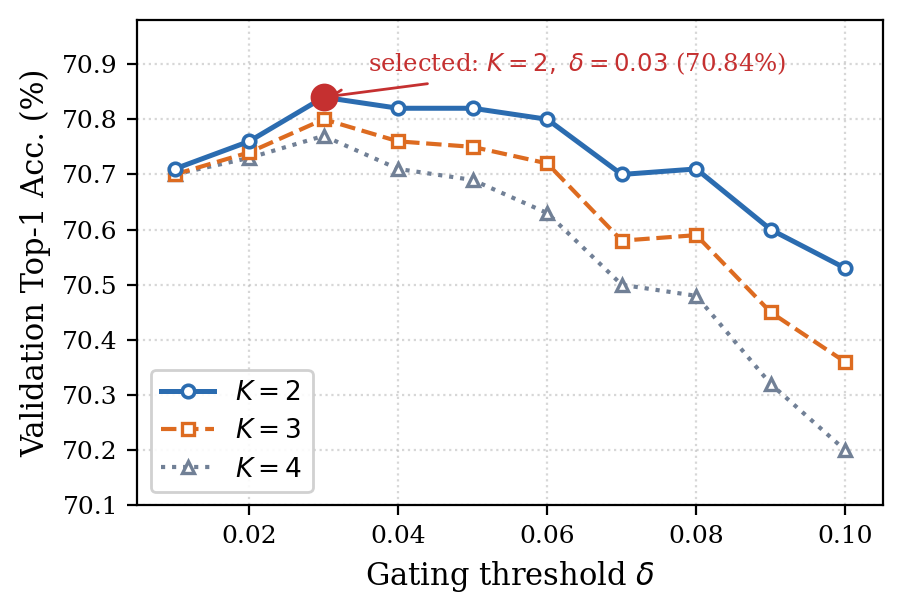}
\caption{Hyperparameter selection for shortlist size $K$ and gating threshold $\delta$ on the validation set (character zero-shot setting). The optimal configuration ($K{=}2, \delta{=}0.03$) is fixed at test time.}
\label{fig:k_delta}
\end{figure}

We select the shortlist size $K$ and gating threshold $\delta$ using a held-out validation set. As illustrated in Figure~\ref{fig:k_delta}, a smaller shortlist consistently yields better accuracy ($K{=}2 > K{=}3 > K{=}4$). Although expanding $K$ improves raw retrieval recall, it also admits negative distractors that trigger re-ranking errors. For $K{=}2$, the model exhibits high robustness to the threshold, with performance fluctuating by less than $0.4\%$ across all evaluated $\delta$. Accordingly, the validation optimums ($K{=}2$ and $\delta{=}0.03$) are selected and fixed for final evaluation.

\section{Conclusion}
\label{sec:conclusion}

In this paper, we propose a coarse-to-fine framework for zero-shot Chinese character recognition. By introducing a spatial-aware IDS representation that injects tree topology and continuous geometric priors, we establish a consistent spatial description across categories, effectively mitigating the seen-to-unseen spatial gap. Our framework coordinates STG-CLIP for high-recall global retrieval and a Radical Verification Module (RVM) for local candidate re-ranking, selectively triggered via a margin-gated rule. Extensive experiments demonstrate that our method achieves state-of-the-art accuracy in character zero-shot scenarios. Our analysis also reveals scaling properties of ViT encoders and outlines limitations in radical zero-shot settings. Explicitly modeling radical geometry and local verification provides a robust paradigm for compositional recognition, and extending the RVM to entirely unseen radicals remains an important direction for future work.

\subsection*{Acknowledgements}

This work was financially supported in part (project number: 112UA10019) by the Co-creation Platform of the Industry Academia Innovation School, NYCU, under the framework of the National Key Fields Industry-University Cooperation and Skilled Personnel Training Act, from the Ministry of Education (MOE) and industry partners in Taiwan. It also supported in part by the National Science and Technology Council, Taiwan, under Grant NSTC-115-2634-F-A49-011-, NSTC-114-2218-E-A49-024-, Grant NSTC-115-2425-H-A49-001, Grant NSTC-114-2622-E-A49-027, Grant NSTC-115-2221-E-A49 -124 -MY3, Grant NSTC-115-2218-E-A49 -017 and in part by the Higher Education Sprout Project of the National Yang Ming Chiao Tung University and the Ministry of Education (MOE), Taiwan. It is also partly supported by E.SUN Financial Holding Co Ltd. (Corresponding author: Ching-Chun Huang.)

%
%
\bibliographystyle{splncs04}
\bibliography{main}
\end{document}